\documentclass[journal]{IEEEtran} 

\usepackage{cite} 

\usepackage{graphicx} 
\DeclareGraphicsExtensions{.pdf} 

\usepackage{url} 

\usepackage[utf8]{inputenc} 
\usepackage{amsmath} 
\usepackage{amsfonts} 

\usepackage{algorithm} 
\usepackage{algpseudocode} 
\newcommand{\algofontsize}{\small} 

\usepackage{tabularx} 
\usepackage{booktabs} 
\usepackage{multirow} 

\usepackage{caption} 
\begin{document}

\title{Swarm Characteristics Classification \\Using Neural Networks}
\author{Donald W. Peltier III,
        Isaac Kaminer,
        Abram Clark,
        Marko Orescanin%
\thanks{Authors are with the Naval Postgraduate School, Monterey, CA, 93943.}%
\thanks{D. Peltier is with the Department of Mechanical \& Aerospace Engineering.}%
\thanks{I. Kaminer is with the Department of Mechanical \& Aerospace Engineering.}%
\thanks{A. Clark is with the Department of Physics.}%
\thanks{M. Orescanin is with the Department of Computer Science.}%
\thanks{Manuscript accepted for publication in IEEE Transactions on Aerospace and Electronic Systems (TAES).}
\thanks{\copyright~2025 IEEE. Personal use of this material is permitted. Permission from IEEE must be obtained for all other uses, including reprinting/republishing this material for advertising or promotional purposes, collecting new collective works for resale or redistribution to servers or lists, or reuse of any copyrighted component of this work in other works.}
\thanks{DOI: https://doi.org/10.1109/TAES.2024.3447615}}%
\date{March 26, 2024}
\maketitle

\begin{abstract}
Understanding the characteristics of swarming autonomous agents is critical for defense and security applications. This article presents a study on using supervised neural network time series classification (NN TSC) to predict key attributes and tactics of swarming autonomous agents for military contexts. Specifically, NN TSC is applied to infer two binary attributes - communication and proportional navigation - which combine to define four mutually exclusive swarm tactics. We identify a gap in literature on using NNs for swarm classification and demonstrate the effectiveness of NN TSC in rapidly deducing intelligence about attacking swarms to inform counter-maneuvers. Through simulated swarm-vs-swarm engagements, we evaluate NN TSC performance in terms of observation window requirements, noise robustness, and scalability to swarm size. Key findings show NNs can predict swarm behaviors with 97\% accuracy using short observation windows of 20 time steps, while also demonstrating graceful degradation down to 80\% accuracy under 50\% noise, as well as excellent scalability to swarm sizes from 10 to 100 agents. These capabilities are promising for real-time decision-making support in defense scenarios by rapidly inferring insights about swarm behavior.
\end{abstract}


\section{Introduction}
\label{sec:intro}
\IEEEPARstart{T}{he} advent of autonomous vehicles in military and civilian sectors presents new challenges and opportunities. Military engagements involving multiple autonomous systems, in particular, have become increasingly significant in various domains, including air, sea, and land operations. The United States Department of Defense (DoD) recognizes the importance of developing strategies to defend against groups of autonomous enemy agents, commonly referred to as swarms \cite{kallenborn_are_2020}. In parallel, the increasing deployment of autonomous agents in civilian applications, notably in transportation, underscores a similar need for advanced tracking and conflict avoidance techniques. This convergence of military and civilian interests highlights the need for sophisticated methods capable of real-time characterization and behavior prediction for scenarios involving numerous autonomous agents.

This need is particularly acute when dealing with swarming tactics, where individual agents collaborate to exhibit complex collective behaviors. Prior studies have explored machine learning applications in classifying trajectories, yet there is a gap in research focusing on real-time inference of individual agent attributes and overall swarm tactics \cite{sami_real-time_2023, gingrass_shape_2021, teatini_movement_2019}.

Central to addressing these challenges is the field of Time Series Classification (TSC), which has emerged as a crucial tool for understanding the temporal dynamics of swarming agents. TSC facilitates the prediction of categorical labels for time-ordered data, a capability vital for interpreting the actions of autonomous swarms. The adoption of machine learning neural network (NN) models like Multi-layer Perceptrons, Convolutional Neural Networks (CNN), Long-Short Term Memory Recurrent Neural Networks (LSTM), and Transformers with attention mechanisms has shown promise in TSC. However, the application of these models specifically to swarm behavior remains an uncharted area of exploration \cite{foumani_deep_2023, vaswani_attention_2017, hauri_group_2022, fawaz_deep_2019, bagnall_great_2016, wang_time_2016, allam_jr_paying_2022, zerveas_transformer-based_2020, jin_time_2022}.

This study introduces an approach using supervised neural network TSC to predict communication and navigation attributes of swarming agents rapidly. By focusing on these attributes, our research aims to discern swarm tactics over brief observation windows, addressing the critical need for real-time analysis in defensive scenarios. Our methodology, tested in simulated swarm-on-swarm engagements, is designed to offer insights into model performance in terms of accuracy, speed, noise tolerance, and scalability, thereby contributing significantly to the understanding of autonomous swarm dynamics.

The key contributions of this research are threefold: (1) it demonstrates the application of NN TSC in predicting the behavior of swarming autonomous agents, (2) it evaluates the performance of various neural network models in this context, and (3) it presents real-time swarm intelligence gathering that balances model accuracy with operational constraints like speed and noise tolerance.

\section{Methodology}
\subsection{Scenario \& Notation}
\label{sec:scenario}
\begin{figure}[h]
    \centering
    \includegraphics[width=1\linewidth]{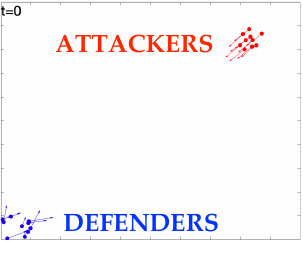}
    \caption{Simulation example showing velocity vectors at initialization time = 0.}
    \label{fig:sim_example}
\end{figure}

Swarm on swarm simulations were conducted in \texttt{Matlab} to generate data for NN TSC. As shown in \figurename~\ref{fig:sim_example} the simulation scenario features an attacking swarm of enemy agents starting in the upper right, shown in red, that is perturbed by a defending swarm of friendly agents starting in the lower left, shown in blue. 

\begin{figure}[h]
    \centering
    \includegraphics[width=1\linewidth]{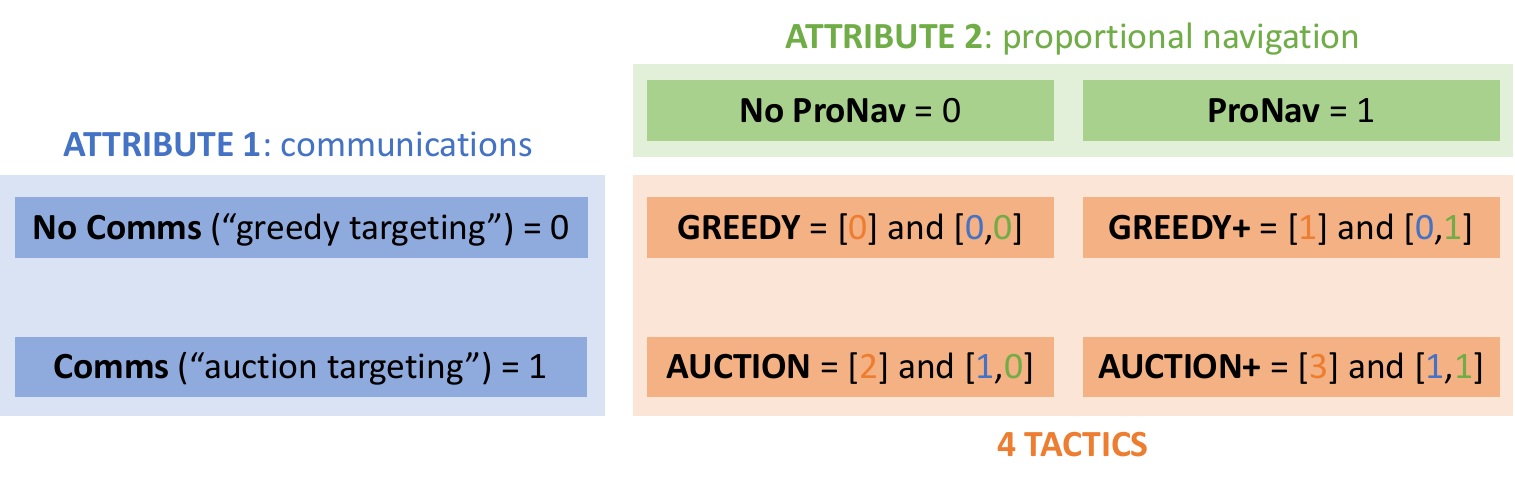}
    \caption{Two binary attacker attributes combine to create four mutually exclusive swarm tactics.}
    \label{fig:comb}
\end{figure}

In this scenario, the defenders are weaponless and aim to motivate the attacking swarm to maneuver, thereby revealing their characteristics. Conversely, the attacking swarm employs one of four tactics. Each tactic, a combination of two binary attributes, is represented in \figurename~\ref{fig:comb}, with each attribute being either active (1) or inactive (0). Tactics are numerically labeled (0, 1, 2, or 3), and also correlate to a specific attribute combination ([0,0], [0,1], [1,0], or [1,1]). During each simulation, all attacker's attributes and tactics are identical and remain constant; there is no switching of tactics. 

Our goal is to classify these tactics and attributes using supervised NN TSC. From a computational complexity perspective, the addition of each binary attribute exponentially increases the number of tactics ($\text{Tactics} = 2^{\text{Attributes}}$), adding to the classification complexity. Thus, in this foundational study, we focus on two pivotal attributes: communication and proportional navigation \cite{siouris_missile_2004}. These attributes were identified as key performance indicators in prior research, including the Aerial Combat Swarms competition \cite{chung_50_2013, buettner_field_2017} and the Service Academies Swarm Challenge \cite{dawkins_deployment_2018}.

Real-world tactics are likely to be significantly more complex, involving more than two attributes, and attackers that may shift tactics over time or employ different tactics within subgroups. The simplicity of the current work is intentional, chosen to demonstrate the feasibility of this application without delving into complex details. This provides a general framework applicable to a multitude of increasingly complex future swarm classification ventures. Potential solutions for relaxed assumptions regarding tactic mixing and switching are discussed in the future work section \ref{sec:future work}, such as applying Group Activity Recognition, sequence output models, and vector output models using sequential classifications from very short time windows.

For this investigation, we consider a two-dimensional (2D) battlespace. The number of attackers and defenders are denoted by \(N_{A}\) and \(N_{D}\), respectively. The positions and velocities of each agent in 2D space are characterized by their \(x\) and \(y\) components, \(P_{x}\), \(P_{y}\), \(V_{x}\), and \(V_{y}\). Each agent is indicated by a subscript: $i$ for attackers and $j$ for defenders. The progression of the scenario is parameterized by time, $t$. Therefore, an attacker's position and velocity at any time $t$ can be expressed as \([P_x(t)_i, P_y(t)_i, V_x(t)_i, V_y(t)_i]\). These time series data serve as the multivariate input for training and evaluating the machine learning models. Specifically, we use only the attacker positions and velocities as input data.

It is immediately apparent that using real-world data would be the most valid approach to support the current findings. However, open-source real-world swarm data could not be found, and likely does not exist, for the variables under discussion (attributes, swarm sizes, noise levels) or at the scale required for NN training and evaluation (i.e., the number of engagements). Data from the Service Academies Swarm Challenge was retrieved, cleaned, and reviewed, but due to the limited number of engagements (approximately 10) and the inherent complexities of real-world swarm-on-swarm exercises (such as unexpected vehicle failures and communication issues), the data recorded held little value within the context of this study. Fortunately, at the macro kinematic level, using simulated data based on Newtonian dynamics provides an accurate approximation of real-world data.

The following assumptions are made:
\begin{enumerate}
    \item All attackers use identical tactics/attributes.
    \item No tactics/attributes switching. 
    \item All attacker full trajectories are known.
    \item If trajectories overlap, agents are distinguishable; no shuffling of data order.
    \item No collision if agents overlap; collision avoided using third-dimension (3D) separation.
\end{enumerate}

For the initial study, the defensive strategy is randomized. Each defensive agent has a unique, randomly assigned constant velocity $V_j$ and heading $\theta_j$, constrained within specific bounds. The defender velocity $V_j$ is kept between 5-40\% of the attacker speed, providing the attackers a speed advantage to prevent endless chases and shorten engagement time. Additionally, the defender heading $\theta_j=\text{Uniform}(0, \pi/2)$ is bounded between North and East to expedite swarm interaction. The initial position for each defender is set near the origin with a spread parameter $S_D=\text{Uniform}(0, 5)$, influencing initial dispersion.

In contrast to the fixed trajectories of defenders, attacker dynamics are modeled through Newton's second law. The force on each agent comprises a thrust force directed towards a virtual leader, and a drag-like term limits the attacker's velocity. Attacker $i$ equation of motion is given by \eqref{eqn:eq-of-mot-att}.

\begin{equation}
    m_i \ddot{\mathbf{r}}_i = K_i \frac{\mathbf{r}_i-\mathbf{r}_{vl}^{(i)}}{|\mathbf{r}_i-\mathbf{r}_{vl}^{(i)}|}  - B_i \dot{\mathbf{r}}_i
    \label{eqn:eq-of-mot-att}
\end{equation}

Here, subscripts or superscripts $i$ refer to attacker $i$, dots denote time derivatives, $m$ is the inertial mass, $K$ is a thrust force magnitude, $B$ is damping parameter, $\mathbf{r}$ is position, and $\mathbf{r}_{vl}$ is the position of a virtual leader (aim point). We note that $K/B = 1$ sets the typical velocity scale and $m/B = 10$ sets a time scale for acceleration up to this velocity, indicating that it takes 10 time steps for attackers to reach the maximum velocity. These equations represent a simplified dynamics model, but are sufficient to generate adversarial swarming engagements that are the focus of this paper. 

Equation~\eqref{eqn:eq-of-mot-att} represents a system of $N_A$ uncoupled second-order ordinary differential equations. At each time step all attacker accelerations are calculated using this equation, based on all virtual leader positions, and we then numerically integrate using a modified velocity Verlet algorithm with a time step $\Delta t$, which we choose as sufficiently small that it does not affect our results. Attacker initial velocities and accelerations are zero, and initial positions include a spread parameter $S_A=\text{Uniform}(0, 5)$ influencing random dispersion around a starting point centroid to the Northeast of the defenders at a range of $40$ distance units in both $x$ and $y$ directions.

As introduced earlier the attacking agents followed one of four tactics (Greedy, Greedy+, Auction, Auction+) where each tactic is a binary combination of two attributes (Comms and ProNav), which will be explained and formally presented below.

The proportional navigation (\textit{ProNav}) attribute determines algorithm used to calculate each attacker aim point, or virtual leader $\mathbf{r}^{(i)}_{vl}$. If this attribute is inactive, each attacking agent uses pursuit navigation. For pursuit navigation, current target position is used for $\mathbf{r}^{(i)}_{vl}$, which typically results in a tail-chase for moving targets because the aim point is guaranteed to lag the target at future time.

Conversely, if the ProNav attribute is active, each attacking agent uses a proportional navigation approximation where the constant velocity vector of the targeted defender and the maximum velocity vector of the attacker are used to calculate an intercept point which is assigned to $\mathbf{r}^{(i)}_{vl}$. This intercept point is calculated every time step, such that if an attacker's velocity is not currently at its maximum speed, then $\mathbf{r}^{(i)}_{vl}$ will change with time. As the attacker's velocity approaches its limiting maximum velocity, $\mathbf{r}^{(i)}_{vl}$ changes become very small over time. Regarding notation, the two tactics using ProNav are appended with a plus sign to indicate improvement (Greedy+ and Auction+), as they are more efficient than their pursuit tactic counterpart.

The communication (\textit{Comms}) attribute refers to an agents ability to communicate with other team members, and is considered linked to target assignment technique employed by the attackers. More specifically, it's typical for a swarm of agents that can communicate to conduct target allocation, or auctioning, to ensure each agent is assigned a different target for efficiency. Tactics with the Comms attribute activated are denoted using the prefix \textit{Auction}. However, if communications are not available, the attackers would conduct an inefficient attack where each agent engages the closest target without deconflicting target allocation, resulting in multiple agents targeting the same target. Tactics with the Comms attribute inactive are represented with the prefix \textit{Greedy}.

Targets for Greedy attackers are assigned based purely on attacker-defender pairwise separation, summarized by Algorithm \ref{alg:greedy}. Each defender is assigned as target for the closest attacker at that instant in time. This process is repeated such that each defender is targeted by the closest attacker to its position.

\begin{algorithm}
\algofontsize 
\caption{GREEDY Targeting Algorithm}
\label{alg:greedy}
\begin{algorithmic}[1]
\Require list of defenders, list of attackers
\Ensure assignments of attackers to defenders
\For{each timeStep}
    \For{each defender in defenders}
        \State Find closest attacker to defender
        \State Assign this attacker to defender
    \EndFor
\EndFor
\end{algorithmic}
\end{algorithm}

Conversely, targets for attackers using Auction are chosen globally according to Algorithm \ref{alg:auction}. Defenders are assigned as targets to the closest unassigned attacker at that instant in time. This process is repeated such that each defender is targeted by the attacker closest to its position from the pool of unassigned attackers. In the case where there are more attackers than defenders, leftover unassigned attackers target the closest defender. Thus, multiple attackers can target the same defender, but only in the case where there are more attackers than defenders. This highlights an important aspect: after each defender is removed the Auction tactics will increasingly resemble the Greedy tactics, as attackers double-up on defenders. This nuance will be discussed further in the results section. Additionally, we note that such a global algorithm requires perfect communication among all defenders or perfect knowledge of the entire scenario, such that each defender can run the algorithm independently and know what all other defenders would be doing.

\begin{algorithm}
\algofontsize 
\caption{AUCTION Targeting Algorithm}
\label{alg:auction}
\begin{algorithmic}[1]
\Require list of defenders, list of attackers
\Ensure assignments of attackers to defenders
\For{each timeStep}
    \State Initialize unassignedAttackers to all attackers
    \For{each defender in defenders}
        \State Find closest unassignedAttacker
        \State Assign this attacker to defender
        \State Remove attacker from unassignedAttackers
    \EndFor
    \If{unassignedAttackers is not empty}
        \For{remaining unassignedAttackers}
            \State Find closest defender to this attacker
            \State Assign attacker to the defender
        \EndFor
    \EndIf
\EndFor
\end{algorithmic}
\end{algorithm}

\begin{figure*}[h]
    \centering
    \includegraphics[width=1\linewidth]{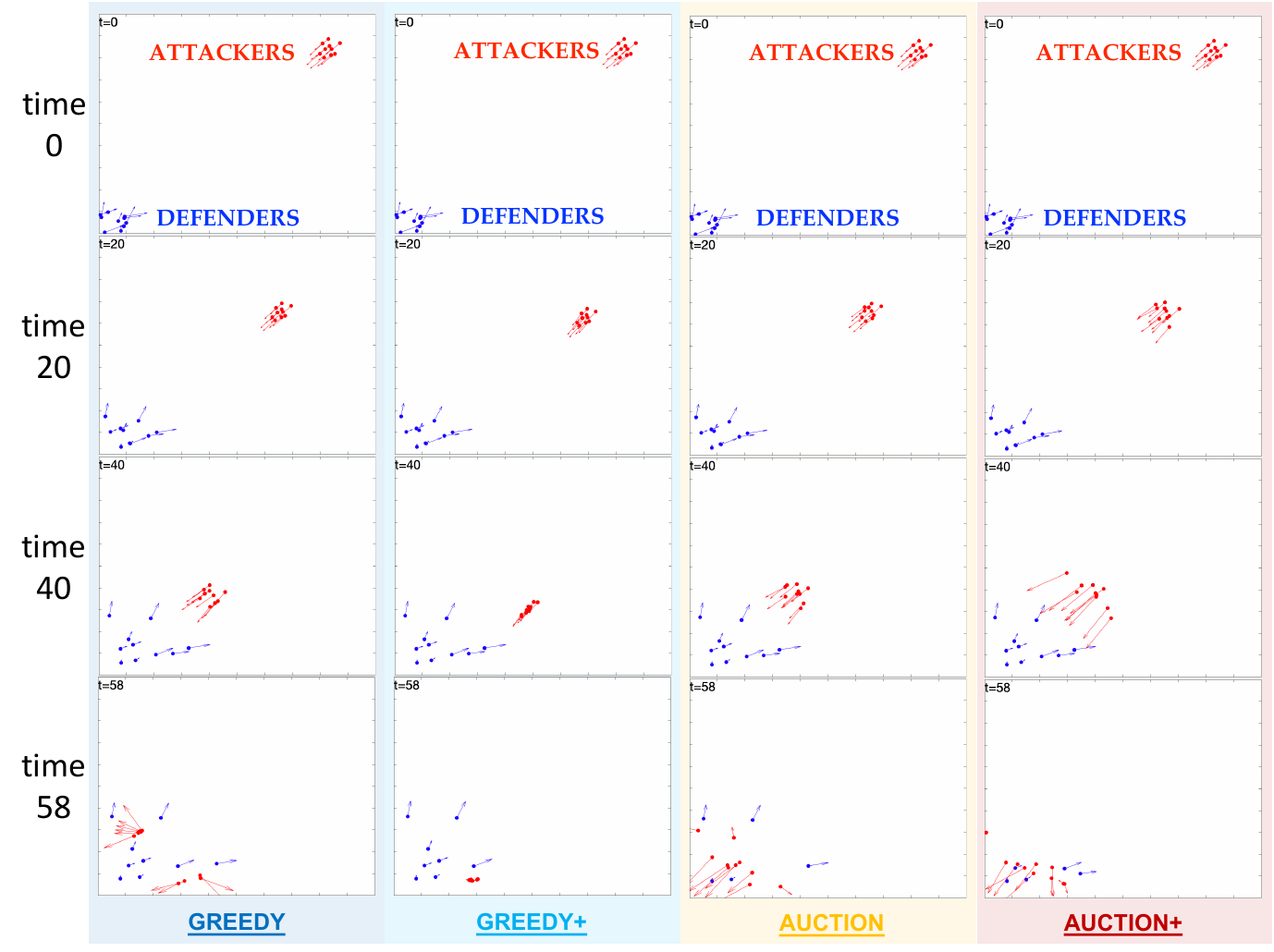}
    \caption{Simulation visualization showing all four tactics (Greedy, Greedy+, Auction, Auction+) with identical initialization at various engagement time steps; agent velocity vectors shown.}
    \label{fig:redblue}
\end{figure*}

For both Auction and Greedy, the targeting process is repeated at every time step, meaning target assignments can (and do) change before the current target is eliminated. This creates a small but important difference between Greedy and Greedy+. As seen in \figurename~\ref{fig:redblue}, specifically comparing Greedy and Greedy+ in the 40 and 58 time step frames, Greedy+ formations increasingly tighten as they travel due to an aim point that shifts less than Greedy aim points, which reduces last minute target shifting (and overshoot).

Unlike the weaponless defenders, the attackers are able to eliminate and remove defenders from the simulation based on a maximum weapons range $R=1$. We track each defenders binary survival $S_j$ at time $t$ according to the attacker-defender pairwise survival function shown in \eqref{eqn:prob-surv-def}. Each simulation instance begins with all defenders alive ($S_j=1$), and then defenders are removed ($S_j=0$) during the course of the engagement based on interactions with enemy agents.

\begin{equation}
    S_j(t) = \prod_i^{N_A} \begin{cases}
        0 &\text{, if }|\mathbf{r}_j-\mathbf{r}_i| < R \\
        1 &\text{, if }|\mathbf{r}_j-\mathbf{r}_i| \geq R
    \end{cases}
    \label{eqn:prob-surv-def}
\end{equation}

Figure~\ref{fig:redblue} shows a simulation example comparing the four tactics over time. It's easily seen that regardless of navigation attribute, both Greedy tactics initially target the closest defender, while the two Auction tactics begin to spread out sooner as they divide the defenders amongst themselves. Note that 20 and 58 time steps were purposefully chosen, as will be discussed in section \ref{sec:exp setup}, while the 40 time step was included to highlight the distinctly different swarm motions.

\subsection{Supervised Neural Network Time Series Classification}
In supervised NN training, the goal is to construct an accurate predictive model by learning from a \textit{dataset} comprised of two tensors (eg. multidimensional vectors): the input data or \textit{features}, and the true output known as \textit{labels}. The process hinges on the NN's ability to predict outputs from the input features. The core objective during training is to minimize the disparity between the predicted outputs and the true labels. This is accomplished by iteratively adjusting the NN parameters to reduce prediction errors. The underlying mathematics of NN training are well documented \cite{geron_hands-machine_2019,goodfellow_deep_2016,ruder_overview_2017,rumelhart_learning_1986} and involves concepts such as gradient descent, backpropagation, and optimization algorithms. These mathematical formulations are crucial for understanding how a NN learns from data and how it updates its parameters in response to the calculated gradients from the cost function. 

In the context of NN TSC, the features are given by time series data. Time series data can be \textit{univariate}, where the dataset is a one-dimensional vector with each element indexed by time. This could represent, for example, a swarm agent's position at each time step. In contrast, \textit{multivariate} time series data involve a tensor with multiple dimensions, where each dimension represents a different variable that changes over time. In this study, multivariate model inputs include attacking agent 2D position and 2D velocity.

A fundamental aspect of NN classification is the differentiation between two types of classification tasks:\\
- \textit{Multiclass} classification deals with problems where each instance is to be classified into one of three or more mutually exclusive classes. In this study classifying the swarm tactic is a multiclass task.\\
- \textit{Multilabel} classification allows for the association of each instance with multiple binary labels simultaneously. In this study classifying the attacking agent attributes is a multilabel task.

Multihead classification refers to a model architecture where the NN has multiple output layers, or ``heads," each responsible for a different prediction tasks. This can be advantageous when there are distinct types of outputs to predict from the same set of features, such as classifying both agent attributes and swarm tactics.

An important facet that Time Series Classification (TSC) adds is the output format \cite{geron_hands-machine_2019}. Many TSC NN models are only capable of producing \textit{vector} output, which is a single prediction at the end of the time series.  However, recurrent models and transformers specifically designed for sequential data can produce \textit{sequence} output, providing predictions for each time step.

To gauge the performance of a NN and guide its training, cost functions and metrics are employed. The \textit{cost function}, also known as loss function, quantifies the difference between the predicted values and the true labels. The goal during training is to minimize this cost function. The multiclass loss function used was \textit{categorical cross entropy} and the multilabel loss function used was \textit{binary cross entropy}.

\textit{Metrics}, on the other hand, are used to evaluate the model's performance according to specific criteria, such as accuracy, precision, recall, or F1 score, which can be more interpretative than raw cost values. \textit{Accuracy}, which is the percentage of correct predictions, is intuitively easy to understand. However, if the possible classifications are not equally represented (ie. more Greedy instances then Auction+ instances), then metrics such as \textit{precision}, \textit{recall}, and their representative combination \textit{F1 score}, provide a more accurate reflection of model performance. Fortunately, in this study all possible labels (tactics and attributes) are equally represented, so accuracy is the primary metric discussed. Additionally, validation loss is also used as a primary metric as it estimates a model's ability to generalize, or predict, on data is has never seen.

For noise impact analysis adjusted accuracy \(Acc_{A}\), which is accuracy shifted for random guess accuracy, is used for a fair comparison, as random accuracy is different for four and two labels. This is based on equally represented random accuracy \(Acc_{R}\) equaling 0.25 for four tactics, or 0.50 for two attributes. Adjusted accuracy \eqref{eq:adj_accuracy} is used to calculate a normalized error rate \eqref{eq:norm_error_rate}, which ranges from 0 for perfect accuracy to 100 for random accuracy. 

\begin{equation}
Acc_A = \text{Accuracy} - Acc_R
\label{eq:adj_accuracy}
\end{equation}

\begin{equation}
\text{Normalized Error Rate} = 1 - \frac{Acc_A}{1-Acc_R}
\label{eq:norm_error_rate}
\end{equation}

\subsection{Experimental Setup}
\label{sec:exp setup}
The experimental approach entailed a structured three-step methodology. Initially, the process began with constructing various neural network models. Next, a baseline dataset was generated by using a sensitivity analysis to understand the dataset's most significant adjustable parameters impact on model performance. Finally, model training involved establishing a baseline for initial comparison, and thereafter tuning the models to minimize validation loss and maximize accuracy.

Machine learning models were developed using \texttt{Python} packages including \texttt{TensorFlow}, \texttt{Scikit Learn}, \texttt{Numpy}, and \texttt{Keras Tune}. Model input data used for this research was synthetically generated using \texttt{Matlab} swarm-on-swarm engagement simulations originally developed at Naval Postgraduate School and only included three tactics \cite{redder_trade-off_2022}. An additional tactic (Greedy+) was created for this research. Lastly, code development was conducted using Naval Postgraduate School high power computing resources\footnote{https://github.com/DWPeltier3/Swarm-NN-TSC}.

\subsubsection{Build Models}
The models considered include those listed in Section \ref{sec:intro}. For models capable of both vector and sequence output (LSTM and TR), both outputs were tested for each model and annotated with a ``V" or ``S" postfix (ie. TR\textbf{V} and TR\textbf{S} for transformer). Additionally, traditional non-NN machine learning classifiers including Random Forest and Logistic Regression were evaluated for comparison. For in-depth model descriptions and diagrams of these well documented NN models, numerous resources are available \cite{geron_hands-machine_2019, wang_time_2016, foumani_deep_2023, vaswani_attention_2017, fawaz_deep_2019}.

Pertinent to this discussion are two benefits of certain model types and outputs. First, all models except the Fully Connected model embed, or transform, the inputs into a preset-dimensional representation, meaning that the exact input dimensions are not important. This allows any input size, accounting for different time series lengths or different number of swarm agents and features. Second, an intriguing potential benefit of sequence output is where tactics switching during operations could be identified.

\begin{figure}[h]
    \centering
    \includegraphics[width=1\linewidth]{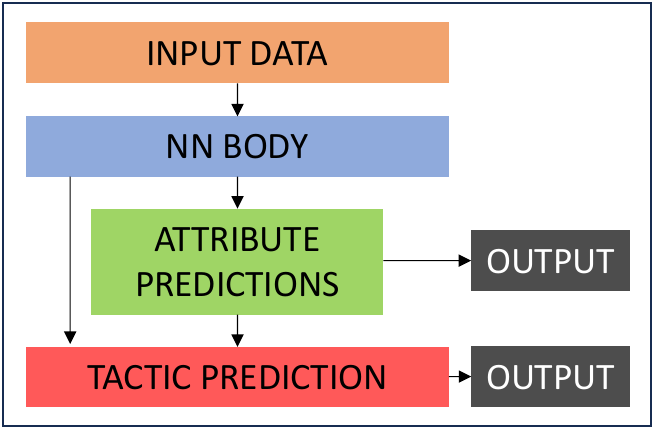}
    \caption{Multihead classifier.}
    \label{fig:mh}
\end{figure}

Estimating that the relationships learned for multiclass and multilabel classification were similar, and therefore NN model weights would be similar, a multihead (MH) classifier was constructed for each NN type. This allowed one model to output both the tactic prediction and attribute predictions simultaneously. Furthermore, as will be discussed in the results section, because the attribute predictions were more accurate than the tactic prediction, attribute prediction loss was weighted heavier (80\%), and the attribute predictions were concatenated as additional features during the final stage of the tactic prediction. This architecture resulted in improved performance, as will be discussed. Figure \ref{fig:mh} shows the basic concept of the MH classifier.

A total of 42 model configurations were examined. These configurations came from five NN architectures capable of vector output (FC, CNN, FCN, LSTMV, and TRV), and two architectures capable of sequence output (LSTMS and TRS), equating to seven architectures examined (5 vector and 2 sequence). Furthermore, each of these seven architectures were evaluated at two different observation window time step lengths, 20 and 58 (``full"), equating to 14 models examined. Lastly, three output types were used: multiclass (MC), multilabel (ML), and multihead (MH = MC \& ML); thus all 14 models were trained on 3 output types, equaling 42 configurations. 

\subsubsection{Generate Dataset}
\begin{figure}[h]
    \centering
    \includegraphics[width=1\linewidth]{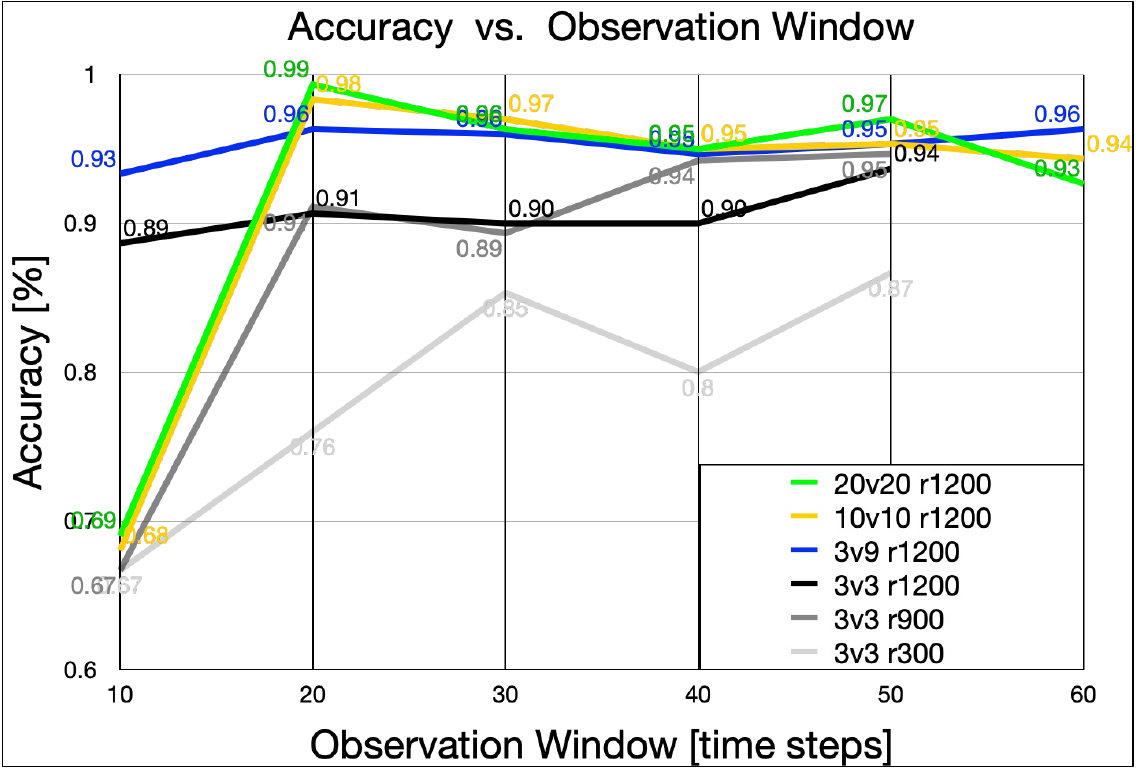}
    \caption{Data generation parameters varied to develop baseline dataset for model comparison.}
    \label{fig:3class}
\end{figure}

A sensitivity analysis was conducted on a typical two hidden layer Fully Connected NN classifier to correlate key data generation parameters to model performance, as shown in \figurename~\ref{fig:3class}, in order to  establish a baseline input dataset for model comparison during training. Parameters varied included number of agents, number of engagements (also known as \textit{instances}), and observation window time steps. Table \ref{tab:base_dataset} shows the finalized baseline dataset parameters. It should be noted that the second best line in \figurename~\ref{fig:3class}, 10v10, was chosen to allow a buffer for possible improvement above 98\% accuracy. Lastly, the noticeable performance spike at 20 time steps led to testing the models at 20 time steps in addition to the full 58 time steps.

\begin{table}[h]
    \renewcommand{\arraystretch}{1.3}
    \caption{Baseline Dataset Parameters}
    \label{tab:base_dataset}
    \centering
    \begin{tabularx}{\columnwidth}{l|l}
    
        \hline
        \textbf{Parameter} & \textbf{Details} \\
        
        \hline
        $N_D$ \text{vs.} $N_A$ & 10 vs. 10 \\
        
        \hline
        Instances & 4800 total = 1200 per tactic \\
        
        \hline
        Time Steps & 58 \\

        \hline
        Training Split & \parbox[t]{\linewidth}{60\% total (80\% training) \\ 2880 total = 720 per tactic} \\

        \hline
        Validation Split & \parbox[t]{\linewidth}{15\% total (20\% training) \\ 720 total = 180 per tactic} \\

        \hline
        Test Split & \parbox[t]{\linewidth}{25\% total \\ 1200 total = 300 per tactic}\\
        
        \hline
    \end{tabularx}
\end{table}

The generation of the dataset involved several steps to ensure features and labels were correctly sized for each specific model input and output. The dataset was initially created using a \texttt{Numpy} array, and then converted into a \texttt{TensorFlow} dataset, which accelerated model training. The features were stored in a 3D tensor format given by [batch, time, feature], and the labels were stored in a similar format for sequence output [batch, time, label]. However, for the vector output, the labels were in a 2D tensor format [batch, label].

\paragraph{Features}
\label{sec:features}
Attacker trajectory data was generated in \texttt{Matlab}, ensuring equal number of engagement instances for each tactic to avoid bias. Next, the data was imported into \texttt{Python} as a \texttt{Numpy} array, which required uniform time lengths. Therefore, all instance time lengths were truncated to the shortest instance time length of 58 steps. Following this, the features were processed following standard neural network data preparation steps including splitting the data into training, validation, and test sets, shuffling for randomness, and scaling the data for normalization, which accelerates gradient decent optimization. The final step was to reshape the data as required. For example, if less than 58 time steps were used (eg: 20 time steps would shorten all instances), or if a Fully Connected  NN was used, which required flattened 1D input for each instance.

\begin{figure}[h]
    \centering
    \includegraphics[width=1\linewidth]{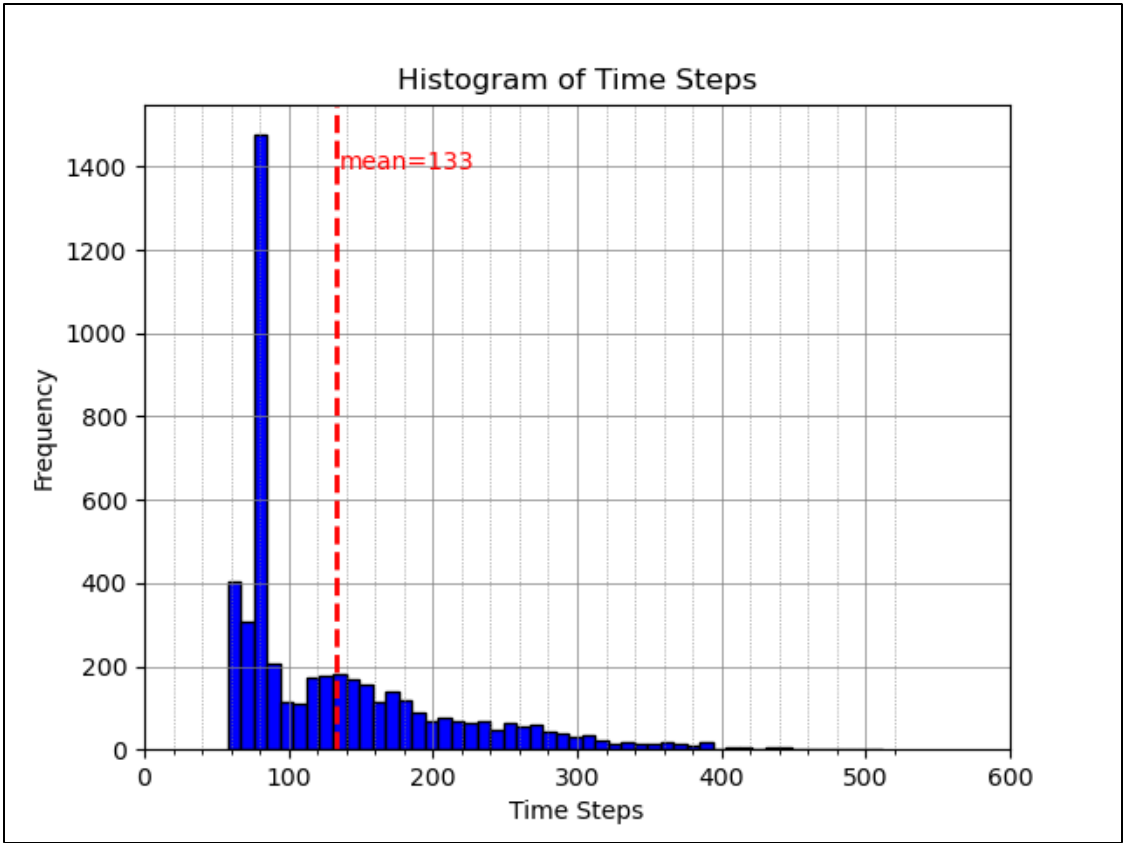}
    \caption{Histogram of time lengths for all generated engagements. All engagements were truncated to the shortest length (58).}
    \label{fig:time_histo}
\end{figure}

An interesting and important aspect of this study is that \textbf{not} all time steps from each engagement were used. As noted above the feature array was uniform across dimensions for simplicity. Though it is possible to import non-uniform (ragged) time series data with varying engagement time lengths, it requires significantly longer NN training times. More importantly, this study wanted to use initial attacker actions during engagement to make the classification.  This was for two reasons. First, the most likely scenario is one where the classification is desired quickly. Second, it's hypothesized that towards the end of engagements all tactics would appear similar as defenders are eliminated and attackers team up on remaining defenders, which would lead to a loss of meaningful information encoded in the NN model. Figure \ref{fig:time_histo} shows the statistical break down of generated engagement time lengths, which highlights that 58 time steps is approximately half the mean engagement time of 133 time steps.

Swarm size scalability analysis required additional datasets with more agents. Datasets were created using 25v25, 50v50, 75v75, and 100v100 agents. For these larger swarm datasets, the number of engagement instances and dataset split percentages shown in Table \ref{tab:base_dataset} were used for consistency. However, the shortest engagement instance varied between 62 and 64 time steps, and the mean engagement length varied between 124 and 144 time steps. When compared to the 10v10 dataset, these ratios of shortest engagement to mean engagement length remained approximately constant providing a fair comparison as swarm size increased.

For noise impact analysis, we augmented our 10v10 dataset by adding zero mean Gaussian white noise to the input features. Recall that the first half of the features are positions, and the second half of the features are velocities, with position and velocity characteristic lengths equal to 40 and 1 respectively. To assess the impact of varying noise levels, we adjusted the noise \textit{scaling factor} from 1 to 50, which represents the percentage of each feature characteristic length that becomes the noise standard deviation. This method allows us to introduce varying levels of noise into our dataset to test the resilience and performance of the neural network models under different noise conditions.

\paragraph{Labels}
Each engagement instance was given a numerical label representing one of the four categorical tactics, and if the output type included multilabel the labels were adjusted to match the attribute combinations, as shown in \figurename~\ref{fig:comb}. For example, the auction tactic label [2] would equate to an attribute label of [1,0], indicating ``Comms" are on, but ``ProNav" is off. Furthermore, if the output was multihead, then both tactic and attribute labels were required, leading to each instance having two labels, one for each output head. Finally, if the output length was sequence, a label was required for each time step. Therefore, each instance label was replicated equal to the number of time steps characterizing the dataset being used.

\begin{figure}[h]
    \centering
    \includegraphics[width=1\linewidth]{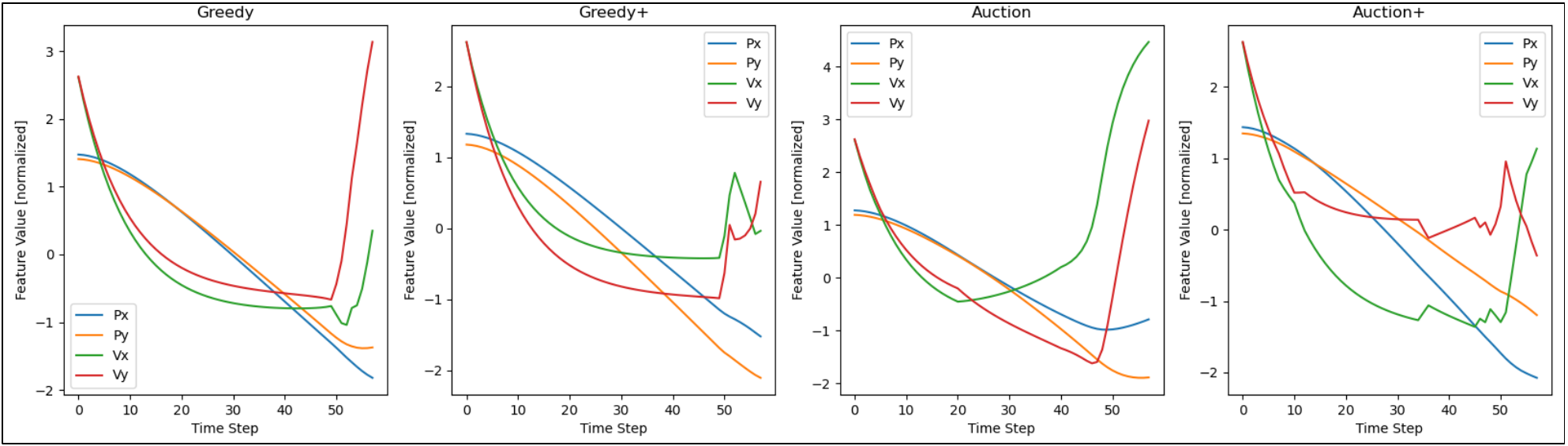}
    \caption{All features (Px,Py,Vx,Vy) versus time for one attacker shown for each tactic.}
    \label{fig:featvis1}
\end{figure}

\paragraph{Data Visualization}
Only attacker position and velocity were used as input data. Feature engineering other potential inputs such as separation distance, velocity gradient, and using defender information were omitted to maintain the simplicity of this baseline research. Figure \ref{fig:featvis1} shows the normalized position and velocity features for one agent during one engagement for each tactic.

\begin{figure}[h]
    \centering
    \includegraphics[width=1\linewidth]{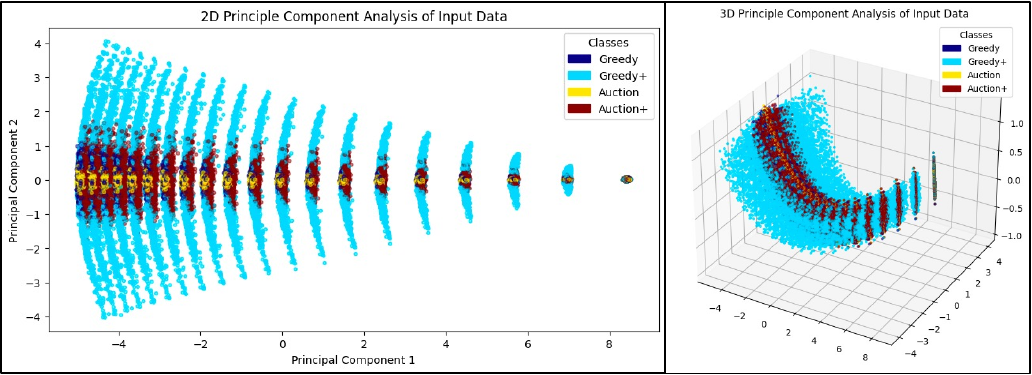}
    \includegraphics[width=1\linewidth]{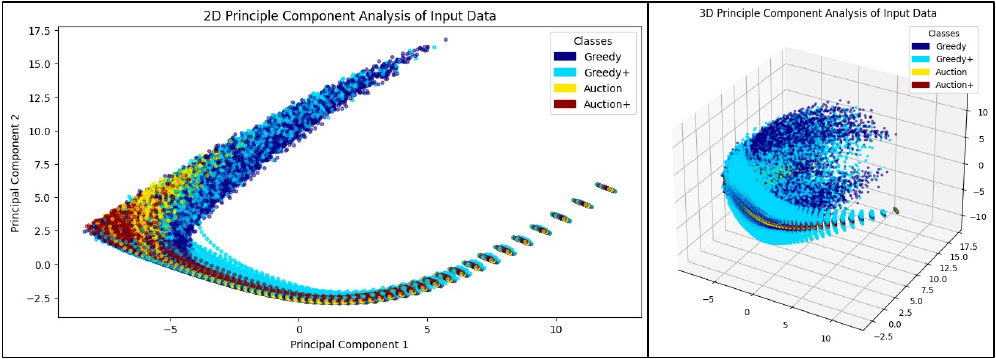}
    \caption{Principle component 2D and 3D visualizations of the high dimensional input features. \underline{Top}: 20 time steps. \underline{Bottom}: 58 time steps.}
    \label{fig:pca}
\end{figure}

Furthermore, \figurename~\ref{fig:pca} shows 2D and 3D Principle Component Analysis (PCA) representations of the 40 input features (ie. 10 agents $\times$ 4 features per agent), allowing visualization of all features across all engagements. Principle Component Analysis representations require 2D input, and were generated using input tensor of shape [batch * time, features], which maintained feature separation. Additionally, this input representation shows time steps along principal component 1 axis, highlighting feature spread at each time step, allowing comparison amongst the four tactics at each time step. Figure \ref{fig:pca} highlights that the ``smarter" tactics (Auction and Auction+), which represent more efficient attack strategies and generally have shorter engagements lengths, are less spread out and more tightly grouped.

\subsubsection{Train Models}
Initial model training used the baseline dataset and hyperparameters (HP) which showed typical performance in previous studies \cite{wang_time_2016}. A small amount of manual tuning was performed to gauge initial impact from certain HP. This resulted in using kernel initialization technique \textit{HeNormal} and using no kernel regularization for all models \cite{geron_hands-machine_2019}. Table~\ref{tab:hp table} list the baseline HP for each model.

\begin{table}[h]
    \renewcommand{\arraystretch}{1.3}
    \caption{Model hyperparameters (HP) varied showing initial ``Baseline" values, allowable tuning window values \{min:step:max\}, and final best performing ``Tuned" values for the multihead full time window configurations.}
    \label{tab:hp table}
    \centering
    \begin{tabularx}{\columnwidth}{l|l|l|l|l}
    
        \hline
        \textbf{Model} & \textbf{HP} & \textbf{Baseline} & \textbf{\{min:step:max\}} & \textbf{Tuned} \\
        
        \hline
        \multirow{3}{*}{FC} & \# Units & 2 & \{1:1:6\} & 3 \\
        & \# Neurons & 100,20 &\{10:10:100\} & 100,100,60 \\
        & Dropout & 0.2 & \{0.2:0.1:0.5\} & 0.2 \\
        
        \hline
        \multirow{5}{*}{CNN} & \# Filters & 3 & \{1:1:6\} & 4 \\
        & Filter & 64,64,64 & \{32:32:256\} & 64,32,192,96 \\
        & Kernel & 3,3,3 & \{3:2:7\} & 3,3,5,7 \\
        & Pool & 2 & \{2:1:5\} & 3 \\
        & Dropout & 0.4 & \{0:0.1:0.5\} & 0.1 \\
        
        \hline
        \multirow{3}{*}{FCN} & \# Filters & 3 & \{1:1:6\} & 2 \\
        & Filter & 64,128,256 & \{32:32:256\} & 96,32 \\
        & Kernel & 8,5,3 & \{3:2:7\} & 7,5 \\
        
        \hline
        \multirow{2}{*}{LSTMV} & \# Units & 2 & \{1:1:6\} & 1 \\
        & Units & 40,20 & \{10:10:100\} & 120 \\
        
        \hline
        \multirow{2}{*}{LSTMS} & \# Units & 2 & \{1:1:6\} & 1 \\
        & Units & 40,20 & \{10:10:100\} & 150 \\
        
        \hline
        \multirow{5}{*}{TRV} & \# Enc & 4 & \{1:1:4\} & 2 \\
        & \# Heads & 4 & \{1:1:4\} & 4 \\
        & Dim Embed & 128 & \{100:100:600\} & 500 \\
        & Dim FF & 512 & \{400:100:600\} & 400 \\
        & Dropout & 0.2 & \{0:0.1:0.5\} & 0 \\
        
        \hline
        \multirow{5}{*}{TRS} & \# Enc & 4 & \{1:1:4\} & 2 \\
        & \# Heads & 4 & \{1:1:4\} & 4 \\
        & Dim Embed & 128 & \{100:100:600\} & 500 \\
        & Dim FF & 512 & \{400:100:600\} & 600 \\
        & Dropout & 0.2 & \{0:0.1:0.5\} & 0 \\
        \hline
    \end{tabularx}
\end{table}

\paragraph{Hyperparameter Tuning}
\label{sec:hparam tuning}

After baseline training, a hyperparameter optimization search (also known as ``tuning") was conducted to enhance model performance using \texttt{Keras~Tune}. Two types of tuners were used; \textit{Random} and \textit{Hyperband}.  Hyperband is a more advanced tuning algorithm \cite{li_hyperband_2018}, but requires a significant amount of memory. Therefore all models except the transformer were optimized using hyperband tuner; the transformer models were by far the largest configurations which necessitated using random tuner due to hardware memory constraints. Regardless of tuner used, the tuning metric minimized was validation loss. Following tuning, all models were re-trained using their respective optimized hyperparameters. Table \ref{tab:hp table} shows the tuned HP for the multihead full window configurations.

\section{Results \& Analysis}

\subsection{Neural Network Performance}
For baseline comparison, the non-neural network machine learning models achieved the following test set accuracy: Random Forrest 84\% and Logistic Regression 32\%. Comparatively, the neural network (NN) models demonstrated higher accuracy and low loss in predicting both tactic and attributes. 

\begin{figure}[h]
    \centering
    \includegraphics[width=1\linewidth]{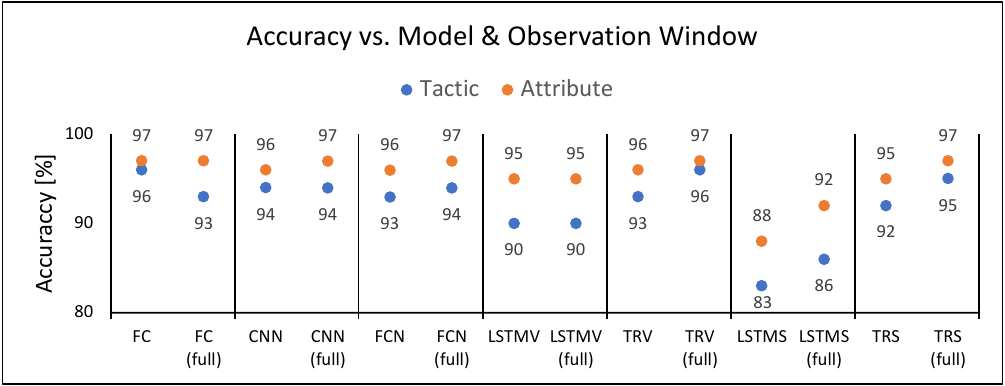}
    \includegraphics[width=1\linewidth]{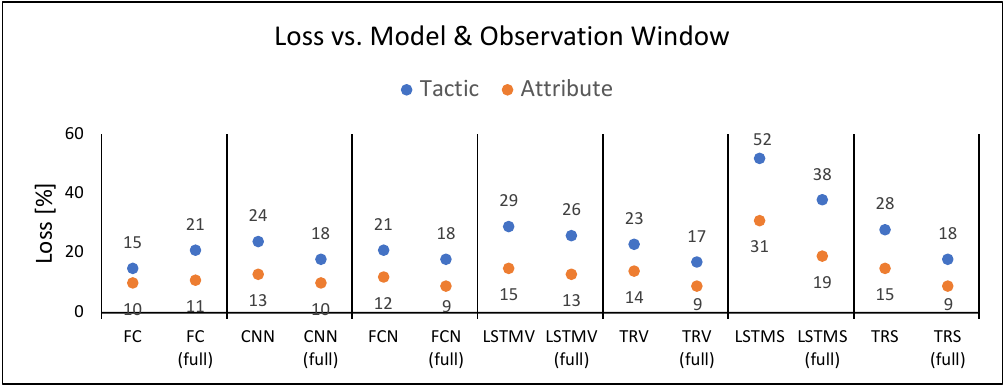}
    \caption{Best test accuracy and validation loss versus model type; each model shows two observation windows, 20 and 58 (full) time steps.}
    \label{fig:best_acc}
\end{figure}

Amoungst the NN models, attribute predictions consistently outperformed tactic predictions, likely attributed to fewer labels and double the training instances (1200 per tactic, amounting to 2400 per attribute). Furthermore, not only did the implementation of a multihead (MH) model facilitate simultaneous prediction of tactic and attributes, it also enhanced 11 out of 14 combined model configurations (separately improved 8 tactic configurations and 10 attribute configurations). This improvement was evidenced by almost universally lower validation losses and higher test accuracies. Additionally, tuning efforts improved 28 out of 42 configurations (67\%), underscoring the importance of model optimization. Lastly, between the two sequence output models explored, the transformer-based model TRS significantly outperformed the LSTM-based model LSTMS. All of these points are summarized in \figurename~\ref{fig:best_acc}.

\begin{figure}[h]
    \centering
    \includegraphics[width=1\linewidth]{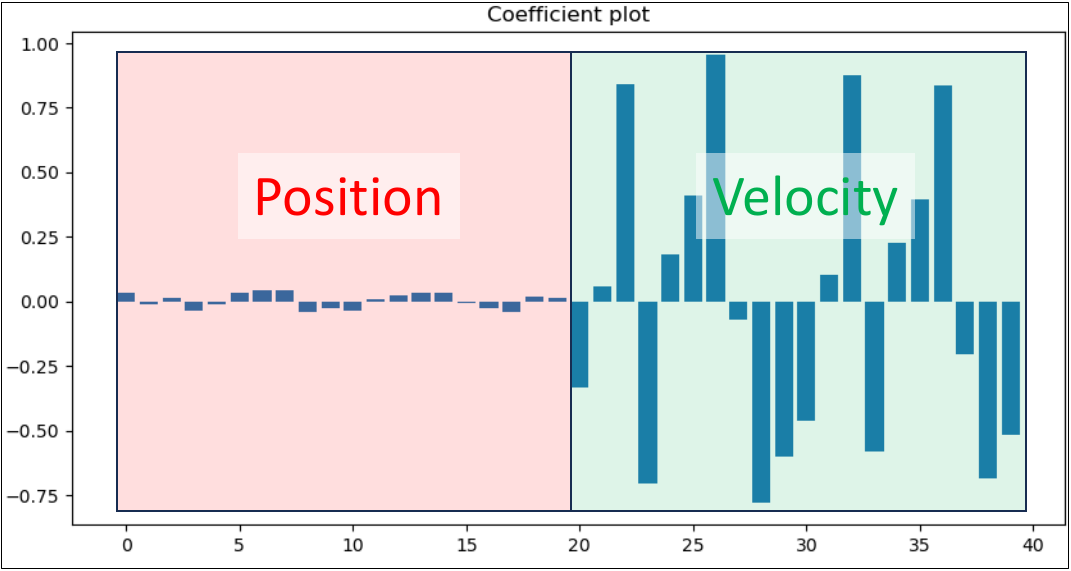}
    \caption{Logistic Regression model feature importance shows position inputs are irrelevant.}
    \label{fig:imp_lr}
\end{figure}

Further illustrations provide insights into NN explainability, which is the ability to describe how a NN is making predictions and how well it is being trained. One example of NN explainability is feature importance, showing which features (inputs) are heavily weighted when models make classification predictions \cite{wang_time_2016}. For instance, Random Forrest and Logistic Regression machine learning models provide feature coefficients used to make predictions, as shown in \figurename~\ref{fig:imp_lr}. Larger magnitude coefficients equate to greater feature importance, indicating that velocity is a more pivotal feature compared to position. This was a key observation; the significant role of velocity as a feature, in contrast to the position. In fact, when position inputs were removed, NN training smoothed and performance showed no degradation. This pattern reinforces the strong correlation between attributes being classified and features provided. It's intuitive that attributes used in this study would be correlated to velocity vectors. This was further highlighted by a Class Activation Map (CAM) analysis conducted to explain the temporal importance on model classification, showing which time steps weighted more heavily when making predictions. The CAM showed a strong connection between attacking swarm total acceleration magnitude changes (ie. velocity vector changes) and time step importance.

\begin{figure}[h]
    \centering
    \includegraphics[width=1\linewidth]{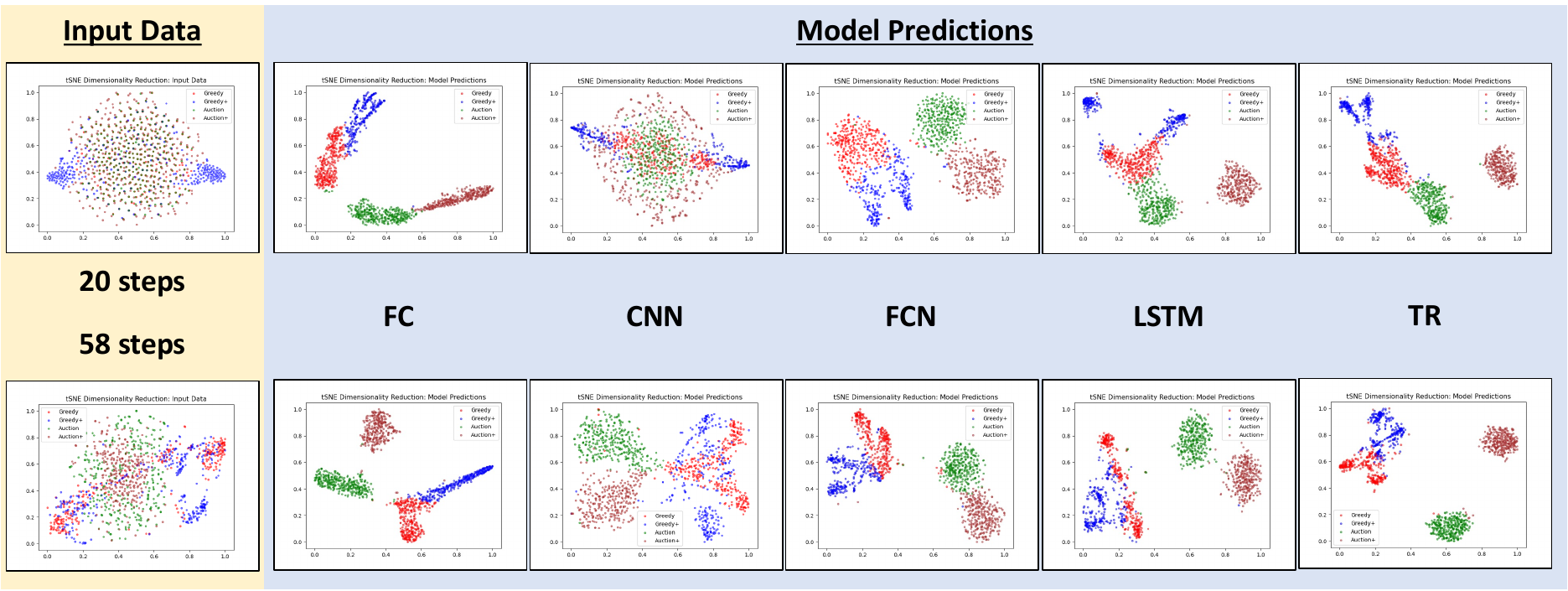}
    \caption{tSNE: disordered input data becomes ordered output, indicating model learning.}
    \label{fig:tsne}
\end{figure}

Another example of NN explainability is the dimensionality reduction technique t-distributed Stochastic Neighbor Embedding (t-SNE) which is a statistical method for visualizing high-dimensional data by giving each datapoint a location in a two or three-dimensional mapping. The tSNE visualization in \figurename~\ref{fig:tsne} effectively demonstrates the transformation from disordered input data to ordered output embedding just prior to the model prediction layer, where clear tactic groupings are indicative of model learning.

\subsection{Observation Window Impact}
When evaluating the effect of observation window length, the 58 time step (full) window offered equal or improved performance for all models except the Fully Connected. Specifically, using 58 time steps improved accuracy for 9 configurations by less than 4\%, as seen in \figurename~\ref{fig:best_acc}. More importantly, all models except FC and LSTM displayed a decrease in validation loss (6-8\%), which is indicative of improved generalization during future inference on unseen data.

An essential point to reiterate is that both the 20 and 58 time step windows are both relatively short when compared to the mean engagement time of 133 time steps, respectively equalling 15\% and 40\% of the mean. Therefore, it is especially interestingly that such a marginal performance increase resulted from training models using 58 versus 20 time steps, which is approximately three times as many time steps. In time critical applications, performing inference three times quicker for a small performance decrease may be a desirable trade off.

\begin{figure}[h]
    \centering
    \includegraphics[width=1\linewidth]{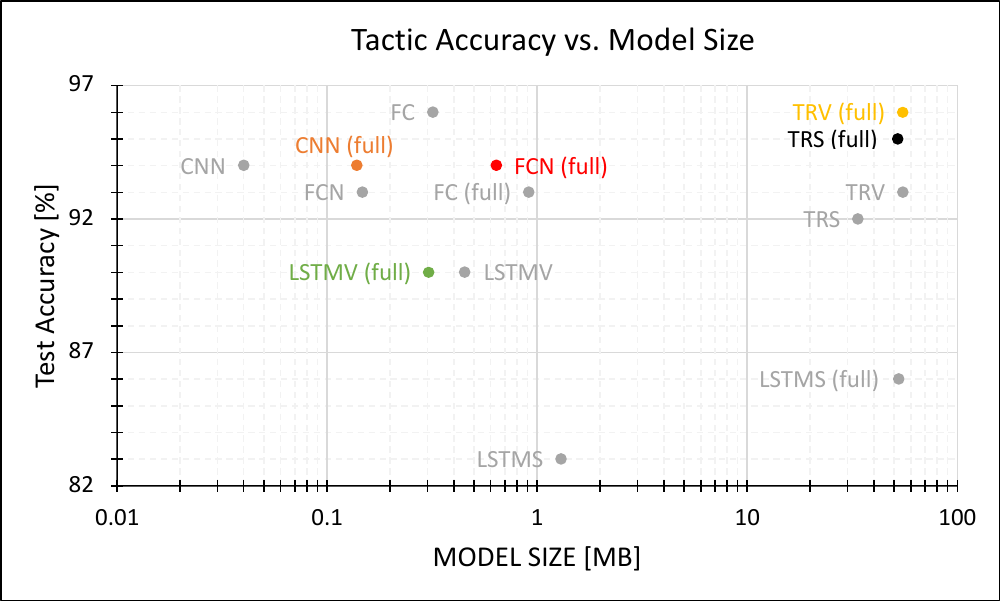}
    \caption{Best accuracy versus model size.}
    \label{fig:best_acc_size}
\end{figure}

As seen in \figurename~\ref{fig:best_acc}, except for the LSTMS architecture all configurations performed approximately the same, with accuracy between 90-97\%. This led to the analysis of model size versus accuracy in order to compare the computational resources and power storage density required to deploy each model on an edge device. A general trend of increasing model size with longer observation windows was observed. Figure \ref{fig:best_acc_size} shows the transformer models were significantly larger (50MB) than other models, while the CNN models were the smallest (0.1MB).

As will be shown shortly in Section \ref{sec:noise}, the five best models were determined to be the full time window configuration for CNN, FCN, LSTMV, TRV, and TRS. Therefore, \figurename~\ref{fig:best_acc_size} has these models highlighted in non-grey color, and they will be the focus for the remaining results analysis.

\subsection{Swarm Size Scalability}
\begin{figure}[h]
    \centering
    \includegraphics[width=1\linewidth]{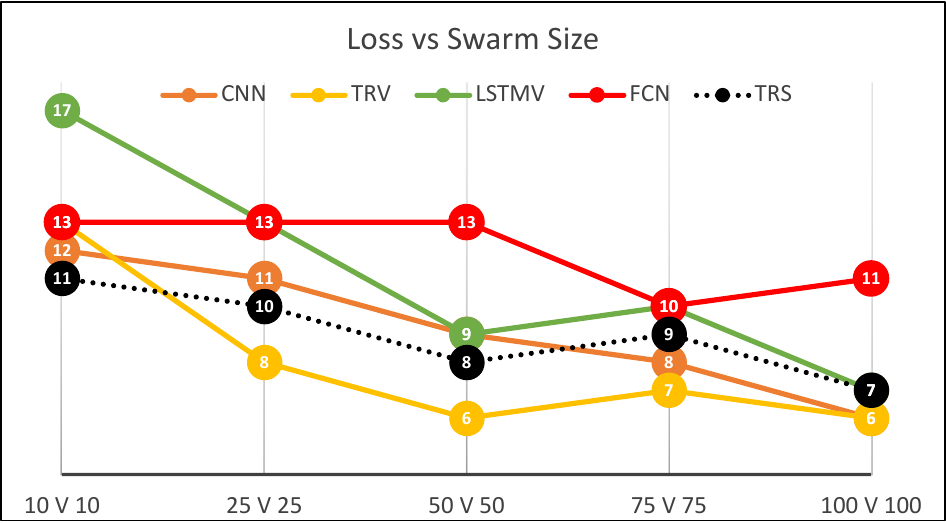}
    \caption{Tactic validation loss vs. swarm size across top five models.}
    \label{fig:swarmsize_loss}
\end{figure}

All models exhibited scalability, which is an ability to be successfully applied to swarms of different size. There is clear improvement for all models as the swarm size increases, as shown by the decreasing validation loss in \figurename~\ref{fig:swarmsize_loss}. The ``best" model was TRV, and the ``worst" was FCN; however, difference was small across all models. However, the best improved was the LSTMV model, indicating that the apparent information deficit previously restricting better performance was overcome by providing more information in the form of increased number of input features; more agents means more information.

\begin{figure}[h]
    \centering
    \includegraphics[width=1\linewidth]{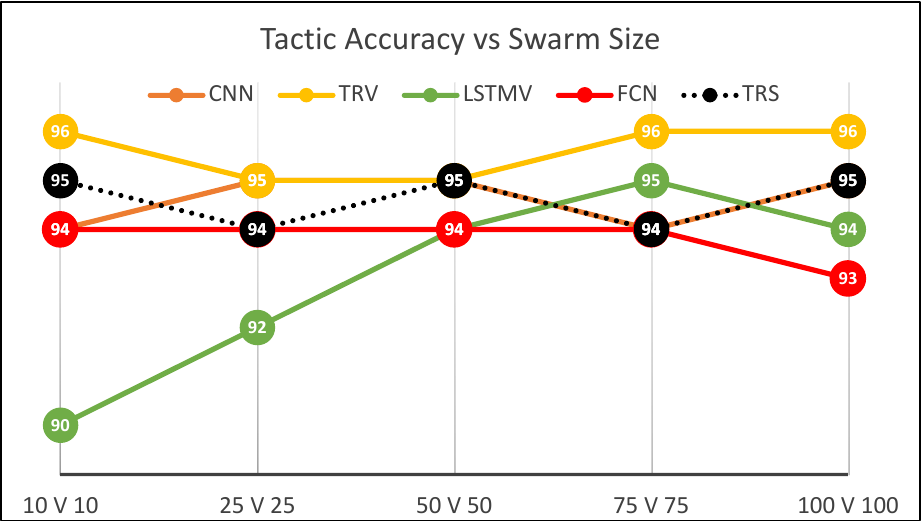}
    \caption{Tactic test accuracy vs. swarm size across top five models.}
    \label{fig:swarmsize_acc}
\end{figure}

Another interesting result is model accuracy remains about constant, or improves, as the swarm size increases, as shown in \figurename~\ref{fig:swarmsize_acc}. This is great, indicating that once a model is performing well, it will continue to perform well as swarm size is increased. This trend implies that one can save time and resources by tuning a model on small swarms knowing that scaling to larger swarms will only improve performance.

\subsection{Noise Robustness}
\label{sec:noise}
\begin{figure}[h]
    \centering
    \includegraphics[width=1\linewidth]{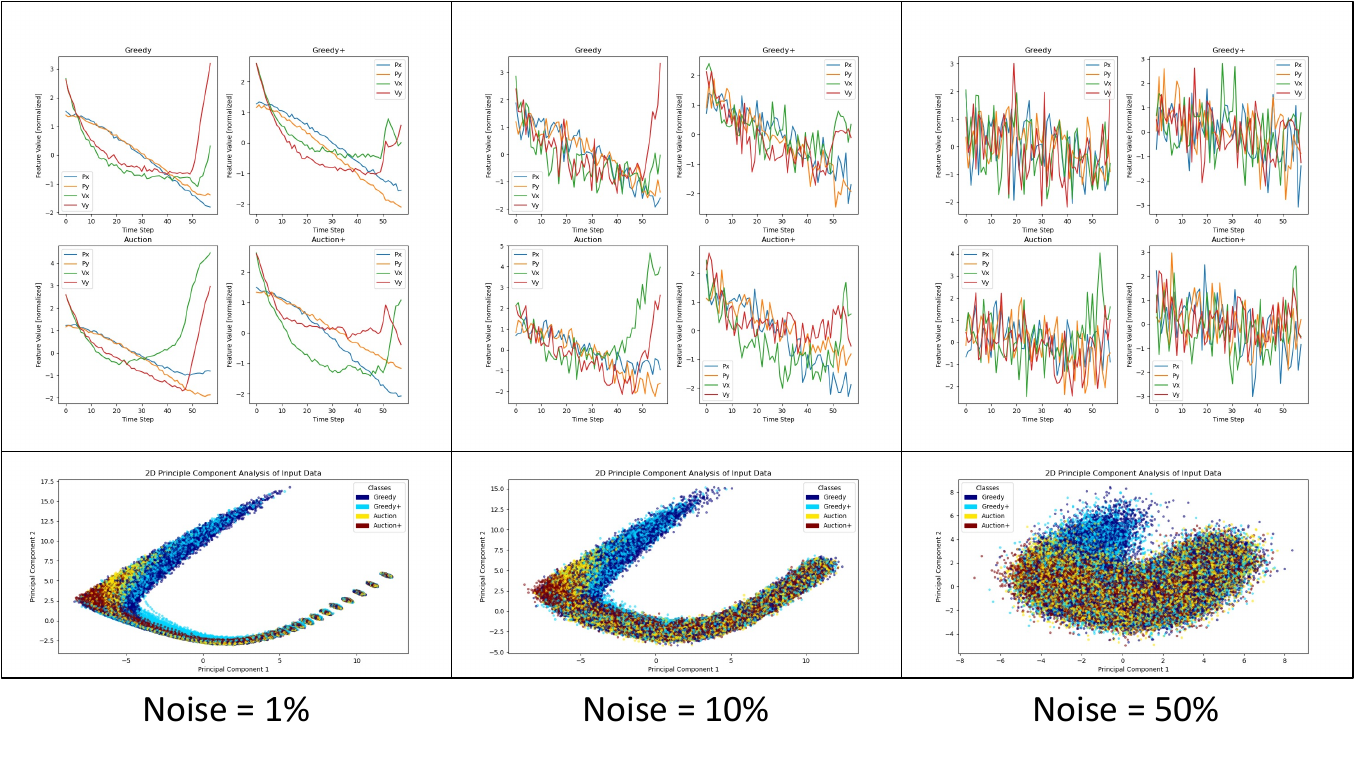}
    \caption{Input data with increasing noise.}
    \label{fig:noisedata}
\end{figure}

Figure \ref{fig:noisedata} shows a canvasing of input features in the presence of increasing noise, as well as the PCA representations which helps visualize the impact of noise on the data.

\begin{figure}[h]
    \centering
    \includegraphics[width=1\linewidth]{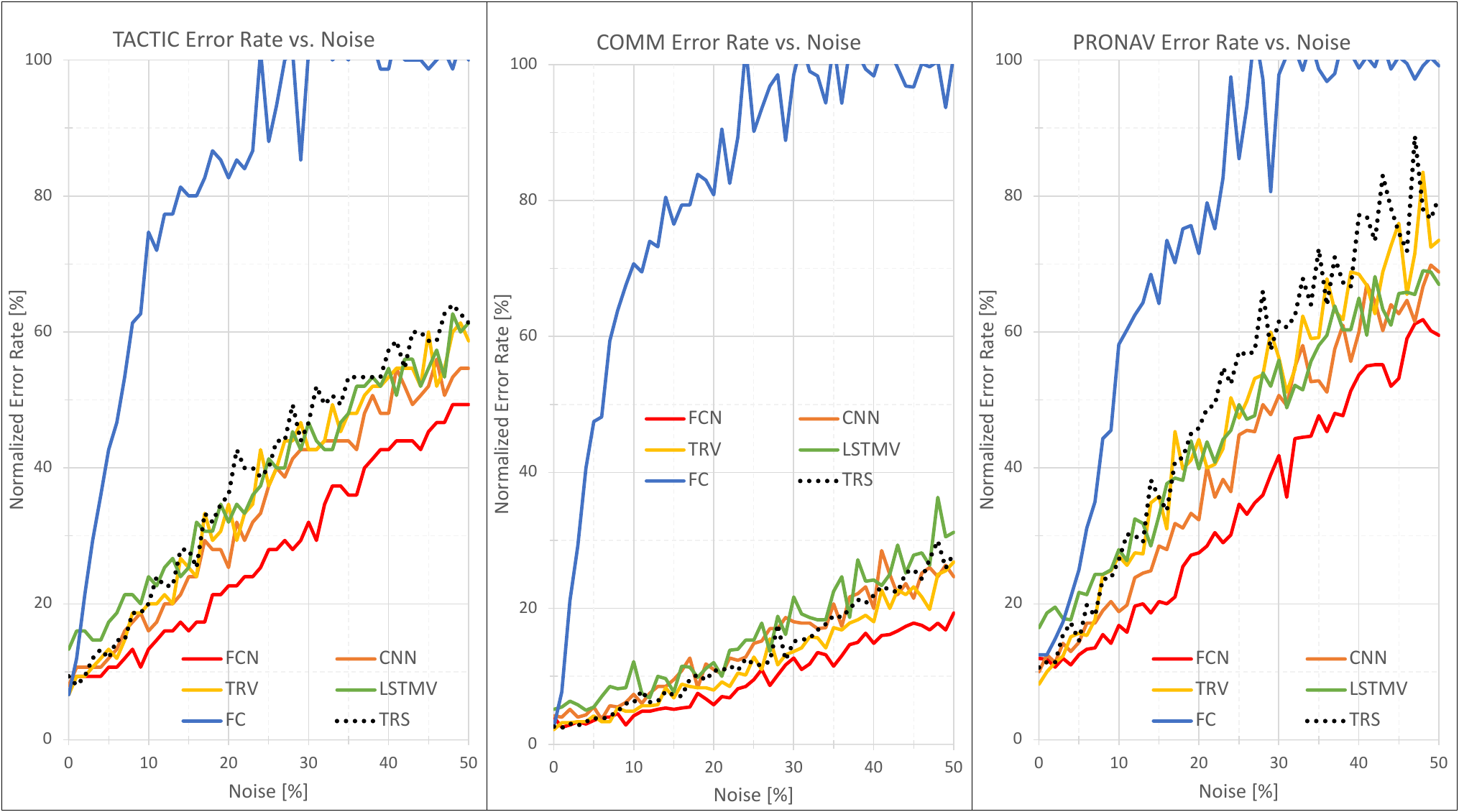}
    \caption{Tactic and attribute prediction error (normalized for random guess) as noise increases.}
    \label{fig:noisecompare}
\end{figure}

The models' performance under varying noise conditions revealed interesting patterns. Noise lightly affected the Comm (targeting) attribute, but significantly affected the ProNav (navigation) attribute, driving tactic error rate similar to ProNav error rate as shown in \figurename~\ref{fig:noisecompare}. This is somewhat intuitive, knowing that the velocity vector was the dominant feature. Specifically, the velocity vector difference between Greedy and Auction targeting was significant, allowing more resistance to noise, whereas the difference between pursuit and ProNav navigation velocity vectors was smaller, making the ProNav attribute more sensitive to noise.

Additionally, the already small velocity vector difference between Pursuit and ProNav decreased further as each attacker closed separation with its target, exacerbating the noise induced prediction degradation. This difficulty in distinguishing navigation method, combined with the observation made in Section \ref{sec:scenario}, where over time as defenders are eliminated the Auction tactics begin to resemble the Greedy tactics, causes a compound effect which manifest as it being most difficult to distinguish the Greedy+ tactic.

Overall, the Fully Connected (FC) model degraded rapidly in the presence of noise, showing it would be an unrealistic choice for real world applications, while the other four models displayed graceful degradation. The Fully Convolutional Network (FCN) had the best robustness to noise, with all other models trailing closely within 5-10\%, offering some ability to rank the top models, but not much.

\subsection{Results Summary}
The top five models (all using full observation window) - CNN, FCN, LSTMV, TRV, TRS - showcased closely matched performance. The key discriminators among these models were their sequence output capabilities and model size, with the transformer models excelling in sequence output, but at the cost of large model size. Conversely, the CNN models were very small, but only capable of vector output. Noise resistance and swarm size scalability offered marginal performance separation between the models, with FCN being most noise resistant, but trailing in scalability. Velocity emerged as the most important feature across all models. Additionally, all of the top models embed, or transform, input information into a latent space of arbitrary dimension, which is important because any input size, along both time and feature dimension, can be accommodated. Lastly, slight to moderate overfitting during training was observed in all models, which was mitigated using early stopping based on minimum validation loss.

\section{Conclusion}

\subsection{Summary of Main Findings}
Throughout the course of this study, several key findings emerged. Neural networks exhibited a strong capability to predict both the underlying characteristic attributes and overall tactics of swarms. Not only were they accurate, but they also achieved this with low loss, leading to robust generalization across various test scenarios. Furthermore, there was a clear trend observed with the volume of data: as the amount of data increased (in terms of more agents leading to more input features, and extended time frames), there was a general improvement in the results. However, a nuanced finding was that while more time data generally contributed to improved results, the increment was often marginal. This suggests an important trade-off in defense scenarios where time is a critical resource. A slightly less accurate classification, which is available faster, might be more practical and desired than waiting for a marginally more accurate one.

\subsection{Potential Applications}
The implications of these findings are profound, particularly in the context of defense. If there's ample training data, this research strongly indicates that essential information about an attacking swarm can be swiftly deduced using neural networks. This rapid inference is invaluable as it can provide crucial intelligence in real-time, leading to better-informed defense decisions.

\subsection{Future Work}
\label{sec:future work}
Future work for enhancing the classification of swarms using neural networks holds a plethora of promising directions, especially considering the constraints and complexities of real-world scenarios. A critical avenue is the focus on utilizing locally observable information, which is vital when sensors have a limited range and cannot rely on global information due to unreliable communications. This constraint presents a more realistic operational environment where the system must act with incomplete knowledge.

Addressing this real-world challenge necessitates the exploration of a more extensive set of attributes and tactics for classification. Realistically, in real-world scenarios there could be an even greater number of tactics, comprised of more binary or non-binary attribute combinations. By canvasing past and current conflicts around the world, as well as evaluating attributes based on their probability of future employment, as list of relevant attributes to be investigated has been developed. With additional attributes identified that could potentially classify an individual agent, the next step is to investigate the relationship between model performance and the increased complexity resulting from a larger number of attribute and tactic labels. Feature engineering will be pivotal in this phase, as the development of more meaningful inputs—whether through additional data or the combination of existing inputs—could significantly enhance model accuracy.

Further, considering the dynamics of engagement, different orientations, starting positions, and swarm spreads must be examined. These factors, coupled with the effect of varying distance scales, raise the question of how to appropriately normalize input data to feed into the neural network. A promising approach could be the use of ragged tensors to incorporate all time steps from each engagement instance into the model. This is a departure from current methods that only utilize the first 15-40\% of time steps for training and prediction. A comprehensive study is warranted to determine if models trained on entire engagements outperform those trained on partial data, even when inference is made using a short time window. Investigating different stages of the engagement, such as the start, middle, or end, could reveal critical insights into optimal prediction windows.

Incorporating tactics switching into the modeling process is another interesting frontier. By simulating tactics changes, such as shifting from pursuit to proportional navigation (ProNav) mid-engagement, researchers can evaluate whether sequence output models or vector output models using sequential classifications from very short time windows can accurately detect these transitions. Following a similar line of thinking, removing the assumption of all attackers using the same tactics and allowing attacking swarms to be a heterogeneous mixture of agents would provide the opportunity to evaluate classification of a more realistic scenario, as well as considering the applications of Group Activity Recognition to possibly parse out the different subgroups of attackers. Alongside this, deliberate perturbations by the defender, as opposed to random, offer a controlled method to study the attackers' responses and how different defender formations and movements impact classification speed and accuracy.

Finally, addressing the effects of input data non-uniformities, such as data gaps from a neutralized attacking agent, scenarios with fewer agents than the model was trained on, or fewer time steps, are crucial. Additionally, the shuffling of input data order, which could arise from crossing trajectories and the difficulty in distinguishing tracks, must be examined to ensure model reliability under various data inconsistencies.

In conclusion, the proposed future work aims to significantly advance the state-of-the-art in swarm classification using neural networks by tackling real-world challenges and enhancing the robustness and accuracy of such systems.

\section*{Acknowledgment}
This work was supported by the Office of Naval Research Science of Autonomy Program and by the Naval Postgraduate School Consortium for Robotics and Unmanned Systems Education and Research.

\bibliographystyle{ieeetr}
\bibliography{0PAPER_classification}

\begin{IEEEbiography}
[{\includegraphics[width=1in,height=1.25in,clip,keepaspectratio]{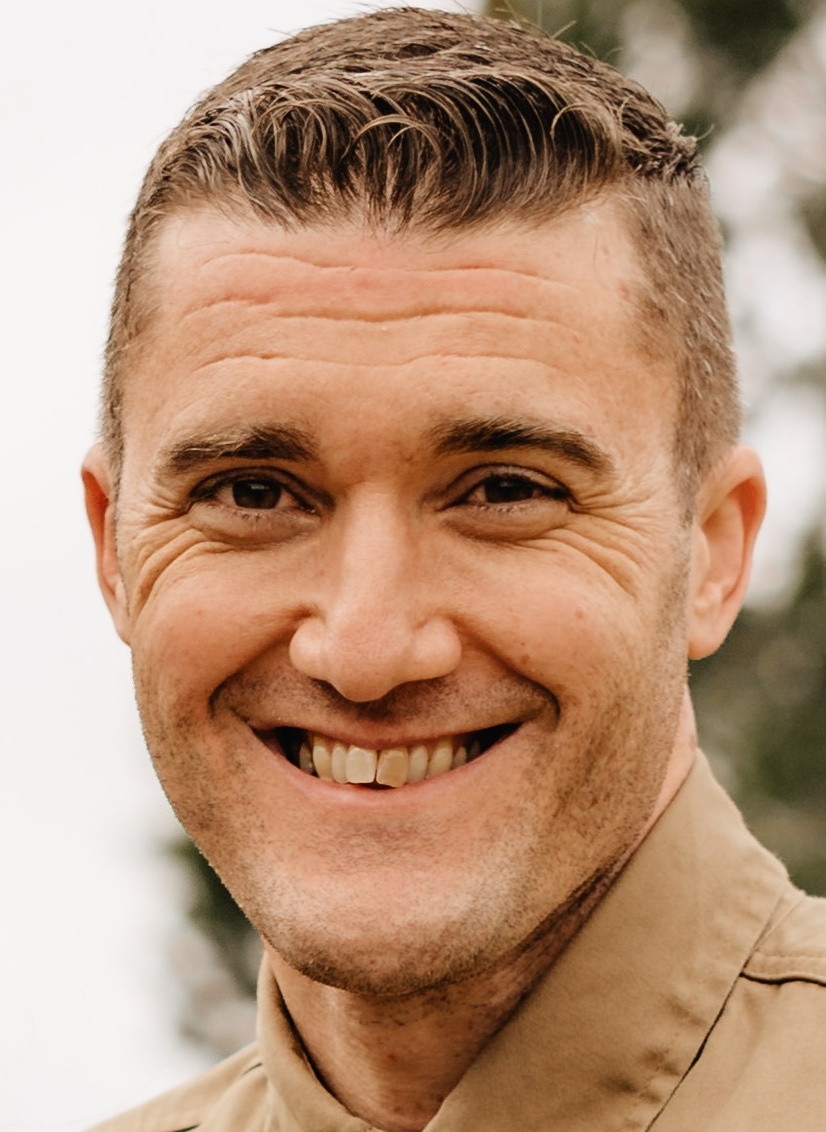}}]
{Donald W. Peltier III}received the B.S. degree in aerospace engineering from The University of Texas at Austin, Austin, TX, USA, in 2006, and the M.S. degree in aeronautical engineering from the Air Force Institute of Technology, Wright-Patterson AFC, OH, USA, in 2007. He is currently working toward the Ph.D. degree in mechanical engineering with a focus in control systems with the Naval Postgraduate School.

His research interests include tasks involving multiple autonomous agents, artificial intelligence assisted control systems, and applications to improve safety and quality of life.
\end{IEEEbiography}

\begin{IEEEbiography}
[{\includegraphics[width=1in,height=1.25in,clip,keepaspectratio]{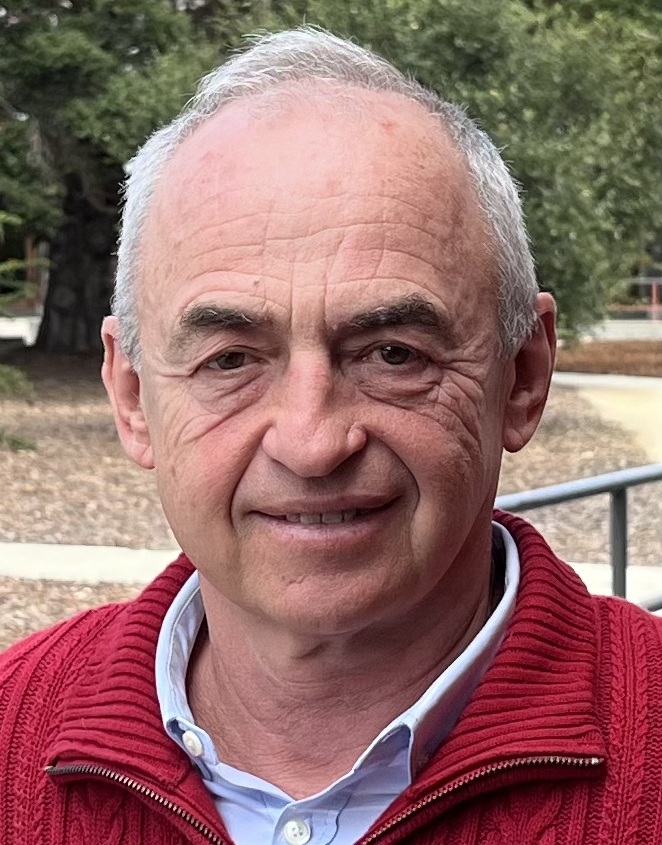}}]
{Isaac Kaminer} received Ph.D. in Electrical Engineering from University of Michigan in 1992. Before that he spent four years working at Boeing Commercial first as a control engineer in 757/767/747-400 Flight Management Computer Group and then as an engineer in Flight Control Research Group. Since 1992 he has been with the Naval Postgraduate School first at the Aeronautics and Astronautics Department and currently at the Department of Mechanical and Aerospace Engineering where he is a Professor. He has a total of over 20 years of experience in development and flight testing of guidance, navigation and control algorithms for both manned and unmanned aircraft. His more recent efforts were focused on development of coordinated control strategies for multiple UAVs and vision based guidance laws for multiple UAVs. Professor Kaminer has co-authored more than two hundred refereed journal and conference publications.
 
He is member of the Institute of Electrical and Electronic Engineers, American Institute of Aeronautics and Astronautics, and Panel Member of the NATO Research Technology Organization, SCI-023 on Unmanned Combat Air Vehicles. He was awarded the NASA Certificate of Recognition for the Creative Development of a Technical Innovation, October 1991; the 1994 NATO Fellowship for Scientific and Technological Exchange, the 1994 Excellence in Research Award, Naval Postgraduate School, the 1995 ASEE/NASA Summer Faculty Fellowship, the 1995 NATO Fellowship for Scientific and Technological Exchange, the 1999 AIAA Outstanding Service Award, the 1999 NPS Menneken Annual Faculty Award for Excellence in Scientific Research, and 2022 IEEE CSS Award for Technical Excellence in Aerospace Control 
\end{IEEEbiography}

\begin{IEEEbiography}
[{\includegraphics[width=1in,height=1.25in,clip,keepaspectratio]{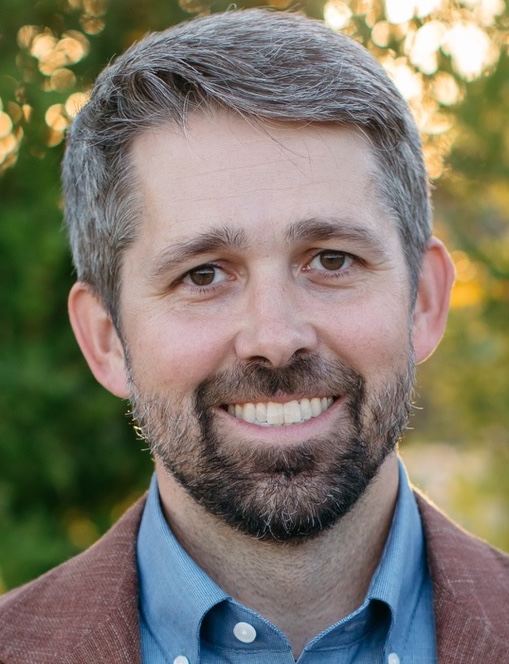}}]
{Abram H. Clark IV} received the B.S. degree in 2006 and the M.S. degree in 2008 in electrical engineering from Texas Tech University, Lubbock, TX, USA, as well as the Ph.D. degree in physics in 2014 from Duke University, Durham, NC, USA.

He was a Postdoctoral Associate from 2014 to 2017 with the Department of Mechanical Engineering and Materials Science, Yale University, New Haven, CT, USA. He is currently an Associate Professor with the Department of Physics, Naval Postgraduate School, where he has been since 2017. His research interests include emergent behavior in soft matter systems (e.g., granular flows) as well as other large, many-body systems (e.g., robot swarms).
\end{IEEEbiography}

\begin{IEEEbiography}
[{\includegraphics[width=1in,height=1.25in,clip,keepaspectratio]{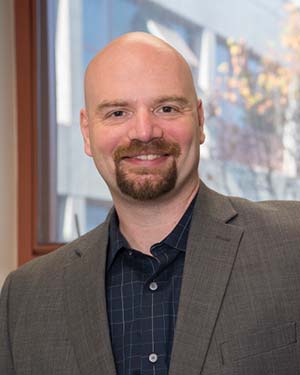}}]
{Marko Orescanin}(Member, IEEE) received a Ph.D. degree in Electrical and Computer Engineering from the University of Illinois Urbana-Champaign, Champaign, IL, USA, in 2010.
Since 2019, he has been an Assistant Professor with the Computer Science Department at the Naval Postgraduate School, Monterey, CA, USA. From 2011 to 2019, he was with Bose Corporation, MA, USA, where he primarily worked on research and advanced development of signal processing and machine learning algorithms for audio and speech enhancement in consumer electronics. He left Bose as a Senior Manager of the AI and Data group with a focus on the consumer electronics business unit. His research interests include signal processing, machine learning, artificial intelligence, Bayesian deep learning, acoustics, passive and active sonar, unmanned vehicles, and remote sensing.
\end{IEEEbiography}

\end{document}